# Inflated Bendable Eversion Cantilever Mechanism with Inner Skeleton for Increased Payload Holding

Tomoya Takahashi, Masahiro Watanabe, Kenjiro Tadakuma, Naoto Saiki, Kazuki Abe
Masashi Konyo and Satoshi Tadokoro, *Tohoku University*

*Abstract*—Inflatable structures used in soft robotics applications exhibit unique characteristics. In particular, the tip-extension structure, which grows from the tip, can grow without friction against the environment. However, these inflatable structures are inferior to rigid mechanisms in terms of their load-bearing capacity. The stiffness of the tip-extension structure can be increased by pressurization, but the structure cannot maintain its curved shape and compliance. In this study, we proposed a mechanism that combines a skeleton structure consisting of multi-joint links with functions to increase rigidity while keeping low pressure and realizing the functions of bending and shape fixation. We devised a design method for rigid articulated links and combined it with a membrane structure that utilizes the advantages of the tip-extension structure. The experimental results show that the payload of the designed structure increases compared to that of the membrane-only structure. The findings of this research can be applied to long robots that can be extended in the air without drooping and to mechanisms that can wrap around the human body.

## I. INTRODUCTION

Rigid mechanisms have features such as high payload capacity and positioning accuracy. Soft robotics, in which a soft, compliant part is in contact with the environment, has many advantages that cannot be realized using conventional rigid mechanisms. One such advantage is that the body is deformable, and hence, precise positioning is not necessary [1].

Among soft robotics, inflatable structures, which are composed of a flexible membrane, can store a large body in a small space [2]. The inflatable structure consists of a flexible membrane that is pressurized in an enclosed space to form a structure that is more rigid than the membrane. In particular, in contrast to the bellow-like extension mechanism shown in Fig. 1 (a-1), the tip-extension structure [3][4] can extend into narrow space without friction against the object (Fig. 1 (a-2)).

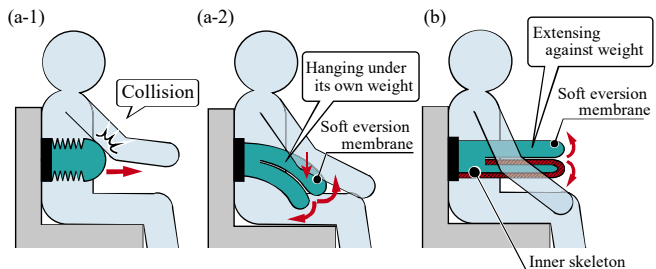

Figure 1. (a-1) Bellows type extension actuator (a-2) Eversion membrane (b) Eversion membrane with skeleton

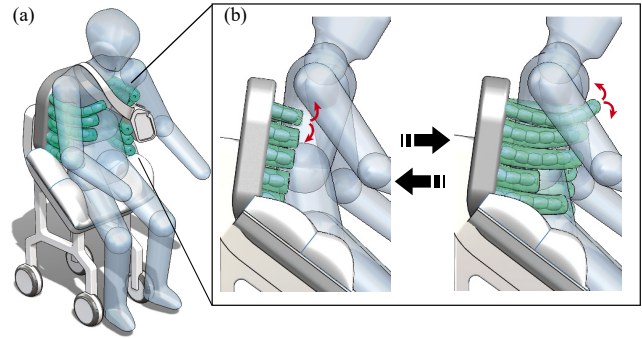

Figure 2. (a) High payload eversion mechanism to support human body (b) Low frictional insertion into under arm by tip extension

Several methods have been proposed to improve the functionality of advanced structures, for example, by installing actuators and devices with rigid bodies inside an eversion robot. Coad et al. [5] showed that the membrane structure causes buckling during retraction; they solved this problem by attaching a tip-winding mechanism. Haggerty et al. [6] proposed a method to expand the workspace, in which a bending motion was added to the winding mechanism. The design of these methods is more complicated than that of methods that use only a membrane structure (in which the size of the space that can be passed through is limited); however, these methods can be applied to a wide range of possible applications of advanced inflatable structures. Our research group previously proposed a tip-extension structure that was driven by a liquid to increase friction against the floor and maintain its shape [7]; a tube-type steering mechanism was inserted inside this structure to select an arbitrary path shape [8]. Such structures have been shown to be effective in search applications in narrow areas owing to their characteristics. Conversely, compliance and movements such as the extension and curvature of the tip are considered to be effective in applications in which it is difficult to apply large forces, such as to human bodies, and wherein an irregularly shaped object is wrapped around the object and supported. For example, for a device that assists in walking, such as that shown in Fig. 2, it is necessary to not only prevent the user from falling off the chair but also to support the user's weight when they are standing up from the chair. (Fig. 1(b), Fig. 2).

Tomoya Takahashi, Masahiro Watanabe, Kenjiro Tadakuma, Naoto Saiki, Kazuki Abe, Masashi Konyo, and Satoshi Tadokoro are with the Graduate school of Information Sciences, Tohoku University, Japan (*Corresponding author: Kenjiro Tadakuma (email: tadakuma@rm.is.tohoku.ac.jp).).

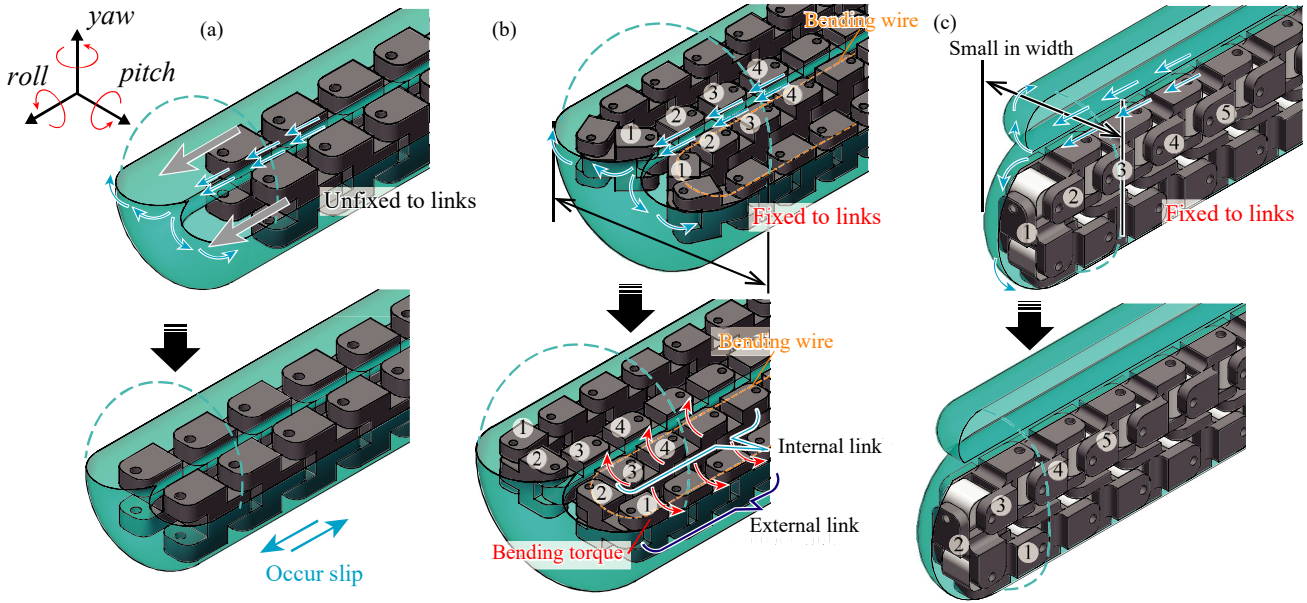

Figure 3. Configuration example of links and membrane (a) Links without folded back separately move to the membrane (b)Folded links with yaw rotation. But it difficult to bend both way by only single wire (c) Proposed structure folded links with pitch rotation

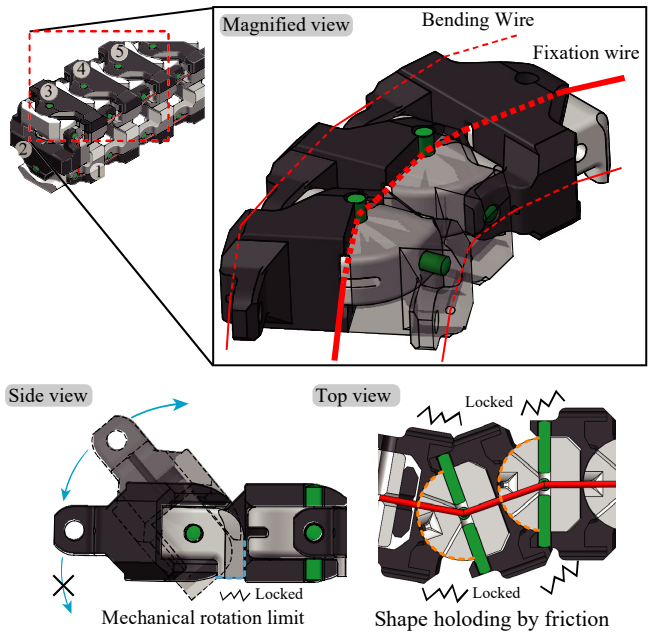

Figure 4. Basic function of articulated links

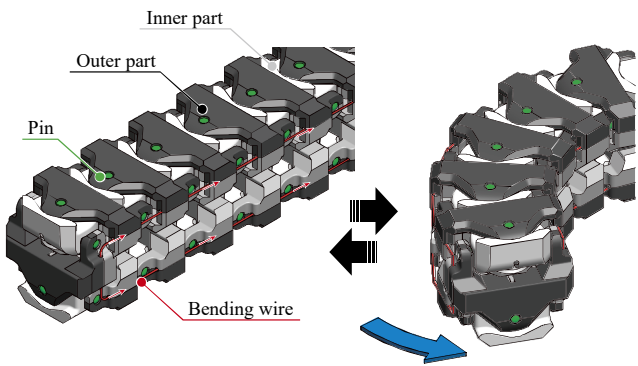

Figure 5. The bending motion by articulated links

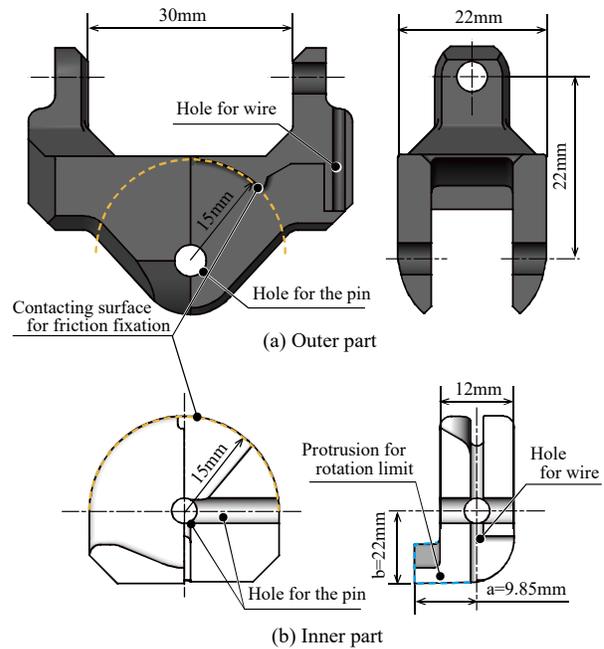

Figure 6. The design of link: (a) Outer part with the hole for bending wire (b) Inner part with protrusion for rotation limit

However, these inflatable structures are inferior to rigid robot arms in terms of load-bearing capacity. For eversion robots with rigid devices or driven by liquid, contact with the ground is required during operation due to the device's own weight. The stiffness of the tip-extension structure can be increased by pressurization [9]. However, an increase in stiffness via pressurization inhibits bending according to the object's shape. Although it is possible to integrate a bending mechanism into an eversion robot [10][11], in any case, the pressure high enough to support a large force makes the structure to low compliance on contact to object. Do et al. [12]. Proposed a method to increase the stiffness of the membrane structure after the stretching motion. In this structure, the membrane itself is shaped like a bag, and multiple sheets are inserted into the bag; the stiffness of the membrane changes according to the friction when the pressure inside the bag is changed. Although this method can increase the payload compared with the conventional single-layer membrane structure, a part of the structure must be soft for the body to be able to bend. In addition, the curvature of the robot is limited by the number and size of the segments. Loh et al. [13]. Nakamura et al. [14]. Proposed a finger structure that uses this tip-extension structure mainly in nursing care; however, the structure had a limited range of motion because it was limited to the tip and assumed contact with the floor.

In this study, we proposed a mechanism that combines a skeleton structure consisting of multi-joint links with functions for increasing rigidity and bending motion (Fig. 1(b), Fig. 2). This mechanism achieves the same level of load-bearing capacity as a robot with a rigid structure; it also combines the softness of the surface and the features of the motion tip-extension structure, enabling an extension of the structure with low sliding with the environment. This structure has a rigid articulated link structure inside the membrane structure, as shown in Fig. 3(c), which acts according to the membrane during the extension motion. The use of articulated links does not interfere with the flexible movement of the membrane and improves the load-bearing capacity. To achieve the grasping motion of heavy objects and the extension and bending motion without contacting the ground, this structure incorporates three types of motion, as shown in Fig. 4 and Fig. 5: withstanding the gravitational load, actively bending in the horizontal direction, and maintaining the curved shape of the body. We devised a design method for rigid articulated links and combined it with a membrane structure that utilizes the advantages of the tip-extension structure. In addition, we experimentally verified the behavioral characteristics of the proposed mechanism using a prototype.

The remainder of the paper is organized as follows. Section II describes the design theory of the articulated links of the proposed tip-extension structure. In Section III, we describe the design of the robot with the articulated links and the fabricated membrane. Section IV describes the experiments using the robot, and Section V discusses the results and describes the limitations of the robot as well as the scope for future improvements. Section VI summarizes the results of this study and concludes the paper.

## II. EVERSION MECHANISM WITH SKELETON

To achieve the functions shown in Fig. 4, the proposed tip-extension structure uses specially designed joints shown in Fig. 6 in the articulated links. This motion is intended for situations where a heavy, soft, undefined shaped object, such as a human body, as shown in Fig. 2, is wrapped and held. In this situation, extension from the tip with low friction is necessary while contacting the object in a situation where it is difficult to contact the bottom of the mechanism to support its own weight. After the structure is curved and deformed to fit the object, it should hold the curved shape so that force need not be applied continuously and the damage to the object can be prevented. The method of changing the inter-joint angle of an articulated link using a tip-mounted mechanism is proposed[15][16]. However, the tip-mounted mechanism increases the deflection because of its own weight, and the size of the tip-extending structure is increased as well. The design of these articulated links and the configuration of the membrane combination are explained as follows.

### A. Basic Configuration of Articulated Links and Soft Membrane Structure

In this study, the articulated links were generally fixed to the membrane structure, and the joints bent in the pitch direction at the tip while extending in accordance with the folding from the tip of the membrane (Fig. 3(c)). The articulated links and the membrane structure can be combined using several methods, but in this study, this configuration was adopted for the following reasons.

First, the articulated links must be fixed to the membrane and must extend while folding back at the tip to ensure that a lifting task can be performed safely. If the articulated links move separately from the membrane without folding back at the tip [17], as shown in Fig. 3(a), the extension of the links must be controlled based on the measured membrane feed because the extension of the eversion structure is half of the membrane feed. In addition, the membrane, which is in contact with the object, may slip against the load-bearing articulated links, and the object may not be held securely.

The folding direction at the tip of the articulated links was assumed to be the pitch direction (gravity direction). For a structure in which the fold in the yaw direction (Fig. 3(b)), the rotation axes of the joints should all be in the same direction. However, it is difficult to bend the entire structure with a single bending operation such as pulling wire. This is because when torque is generated at each joint for the entire bending, the inner and outer links bend in opposite directions. In addition, to the purpose of inserting under the human arm, the width of the structure needs to be small. Therefore, we chased the links has pitch and yaw joint, and the fold was set along the pitch axis direction (Fig. 3(c)). In this structure, the pitch joint cannot be completely free of rotation because it needs to support the load in the direction of gravity. This problem was solved by limiting the range of rotation. As shown in Fig. 4, it can be rotated upward by a maximum of 90° to enable folding, while it does not rotate downward to support weight. These links can constrain the displacement in only one direction, but both upward and downward displacements can be constrained by placing the articulated link in a folded position.

In a conventional eversion robot, which is composed of only a single-layer membrane, buckling occurs during the retracting motion. The stiffness of the proposed link mechanism can prevent this buckling. The tension required for the eversion robot to retract increases in proportion to the pressure, but the stiffness against buckling does not increase significantly with pressure. Hence, it is difficult to prevent buckling using only the membrane structure. In contrast, the proposed mechanism can maintain rigidity as it includes articulated links; thus, the retraction can be realized with a low-pressure retracting tension. In the design of the articulated links, the vertical buckling is geometrically constrained by the angular limit of the links, which prevents buckling. The buckling in the horizontal direction is constrained by the static frictional force between the joints, although the joints are designed to be free.

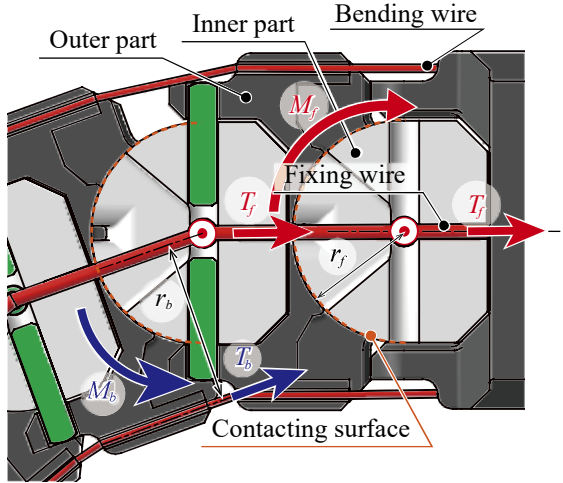

Figure 7. Schematic for bending moment caused by tension of bending wire and link parts

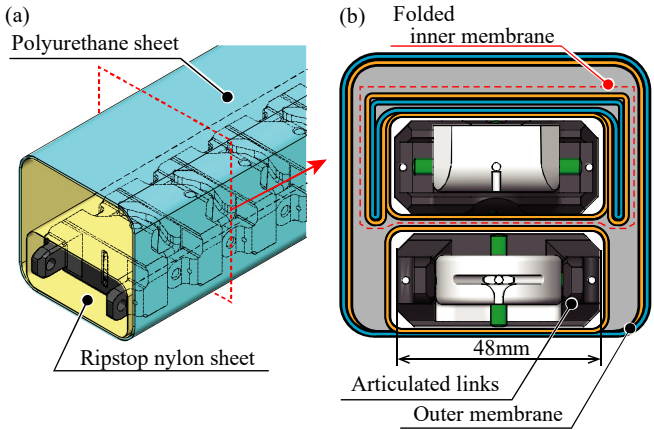

Figure 8. Diagram of the membrane structure
(a) Configuration of layered membrane structure,
(b) Cross-sectional view of red dashed line in (a)

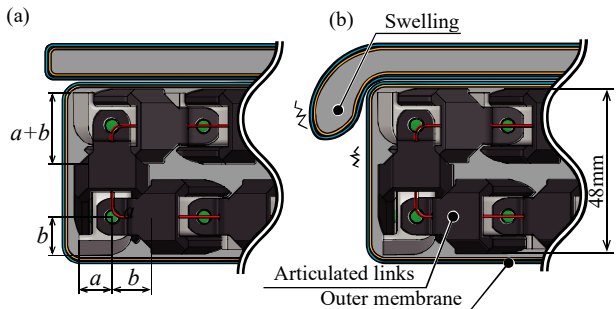

Figure 9 Swelling with the oversize of the membrane

### B. Realization of Active Bending and shape-maintaining Function

An articulated link structure consisting of links with two rotational joints was devised to realize bending and shape-fixing motions (Fig. 4). In this study, we did not use a ball joint; rather, we used a two-axis rotating joint with one degree of freedom because the range of motion was not wide. Thus, the folded part of the link at the tip was large. In addition, the ball joint had a lower rigidity than the rotational link structure with two independent axes; therefore, it was not suitable for the intended application. Therefore, we designed a link based on the universal joint. In addition, only contact area of yaw joint increase to achieve high holding torque by friction.

The bending action was performed by pulling the wires attached to the left and right sides of the articulated links so that the entire link was curved (Fig. 5). A method of attaching a contracting actuator to the surface of the membrane was proposed as the bending mechanism of the tip-extension structure [18][19]. However, this method limits the curvature of the structure owing to its limited contraction rate. In the proposed mechanism, the entire body is curved owing to the articulated links to achieve a larger radius of curvature. As shown in Fig. 7, the bending torque $M_b$ is expressed by the distance between the center of the joint and the wire $r_b$ and wire tension $T_b$ as follows

$$M_b = r_b T_b. \quad (1)$$

The shape of this articulated link could be fixed by pulling a wire through the center of each component, and the link shape could be fixed by friction[20]. Kiryu et al. [21] and Wang et al.[22] proposed an eversion structure to fix the curved shape by friction between membrane structures, but to obtain a larger holding torque, it is better to apply a large force to a rigid structure. Therefore, we adopted a shape-fixing function for articulated links. As shown in Fig. 7, pulling the center wire increases the contact pressure between the parts of the entire link and increases the static friction force at each rotating joint. This constrains the angle of each joint and maintains the shape after bending. The relationship between the wire tension $T_f$ and holding torque $M_f$ is expressed as

$$M_f = \mu r_f T_f. \quad (2)$$

Here, $\mu$ and $r_f$ are friction coefficient and the radius of the contact area, respectively. The expression suggests that a larger holding torque can be achieved by increasing the radius of the contact area.

### C. Combination with Membrane Structure

In the proposed mechanism, only one set of articulated links was inserted into the membrane structure, as shown in Fig. 3(a) and Fig. 8. In addition, the skeleton and the membrane are not fixed in all links, but the membrane and framework are only fixed to the base part and strapped together at the tip. In fixing the membrane and link, to avoid conflicts between the membrane structure and articulated links, we considered the following points.

In this mechanism, only one articulated link, rather than two, was inserted into the membrane, and the configuration was vertically asymmetric. Otherwise, the diameter would be extremely large if two frames were inserted in a vertically symmetrical arrangement, and it would not be possible to utilize the tip-extension structure's advantage of entering narrow spaces. The articulated links and the membrane were fixed in the diameter direction but not completely fixed in the axial direction to ensure a misalignment between the membrane and the articulated links during left-right curvature.

When such a rigid link structure is combined with a soft membrane structure, wrinkles are formed at the tip of the eversion structure, as shown in Fig. 9, during the extension and retraction operations. Furthermore, the membrane structure buckles when the length of the membrane is excessively long in the axial direction. The lower membrane, which is constrained by the articulated link, extends in accordance with the link; in contrast, the upper membrane, which is not constrained by the link, swelling and interferes with the movement if the length is extremely large (Fig. 9). Therefore, the articulated links and the membrane are not fixed in all links but the membrane and links are only fixed to the base part and tied together at the tip at an optimum length.

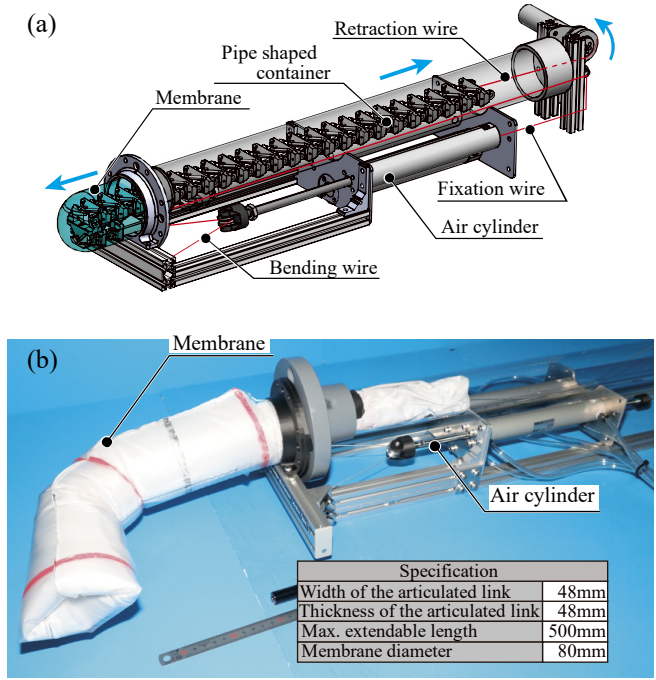

Figure 10. Integration of skeleton and membrane structures and actuators (a) Schematic for prototype, (b) Bent model of prototype

In the articulated links used in this study, the center of rotation for the pitch axial folding motion was on the neutral axis of the link; therefore, slippage might occur at the contact surface between the joint and the membrane during the folding motion at the tip. Hence, the articulated links and the membrane were fixed in the diameter direction rather than in the axial direction. The length of the membrane must be determined by considering this offset $l_{offset}$. This is given by the following equation:

$$l_{offset} = 2a + 2b. \quad (3)$$

where $a$ and $b$ are values determined by the thickness of the link and the distance between the rotation joint, as shown in Fig. 6 and Fig. 9.

III. MECHANICAL DESIGN

A. Design of articulated links

The two types of parts used as components of the articulated links in this mechanism are shown in Fig. 6. The two links are connected by inserting the inner parts into the inner recesses of the outer parts and press-fitting pins from the outside. The two parts have a hole in the center so that the wire can pass through for shape fixing, and the outer part has a hole so that the wire can pass through for left–right bending. The range of rotation was 90° to 0° in the pitch direction and ±30° in the yaw direction. The pitch and yaw axes of the inner part were on the same axis, and the distance between the two axes was 27 mm in the outer part. The articulated link has a total length of about 1000 mm and a mass of 706 g. These specifications were designed not only to support the human torso, as shown in Fig. 1, but also to be able to bend to various curvatures as a mechanism for holding irregularly shaped objects, such as food and animals. The two parts were made with a 3D printer Form3, and Rigid4000(tensile module: 4.1 GPa, ultimate tensile strength: 69MPa) was used as the material.

When the rotation of a joint is constrained by friction, the distance between the joints changes. Therefore, it is not possible to use bearings that completely constrain the direction of rotation. In the proposed mechanism, a larger clearance is provided for the pins of the rotary joint. Thus, when the wire is pulled, the inter-joint distance slightly decreases, and the pins and parts do not contact each other, but the surfaces of the parts contact each other.

B. Design of membrane structure

The membrane structure used in this study consisted of three layers, as shown in Fig. 8. This structure was made of a composite of polyurethane sheets and a slippery nylon material, which were thermally welded together to form a bond. The inner pouch is an area for inserting the articulated links, which allows some movement in the axial direction but restricts movement in the diameter direction. The diameter of the membrane was set at 80 mm to accommodate the articulated links.

C. Integration of membrane structure and articulated links

The membrane structure and articulated links actuated using an air cylinder and motor were integrated, as shown in Fig. 10. The membrane structure and the articulated link structure were placed inside a sealed cylindrical container and were extended by air pressure. The wire attached to the tip of the articulated links was wound by a motor, and tension was applied to the wire for retraction. An air cylinder was attached to the bottom of the pipe-shaped container to pull the wires threaded on both sides of the articulated links. The flexible and shape-fixing wires in the center of the articulated links were connected to the air cylinder through the backside.

IV. EXPERIMENT

A. Measurement of holding torque of links

As described in Section II, in the articulated link mechanism, the joint angle in the yaw direction is fixed by pulling the central wire. The holding torque was measured experimentally. For the experiment, we used an experimental setup with a set of inner and outer parts, as shown in Fig. 11. This experimental setup was attached to a tensile testing machine, and the joints were fixed using friction by pulling a wire with an air cylinder. After fixation, a tensile tester was used to generate torque at the joint by pushing it from above, and the torque applied when the joint slid out was measured. The wire tension was varied from 50N to 150N and the holding torque was measured. The joint angles were set at 0°, 15°, and 30°. To maintain the positional relationship between the center of rotation of the link and the action point of the tensile testing machine, differently designed parts were used at each angle, as shown in Fig. 11. All parts had the same design for the contact surface. This experiment focused on the holding torque caused by friction, so Measurements were taken five times, and the average of the measurements was used for the results.

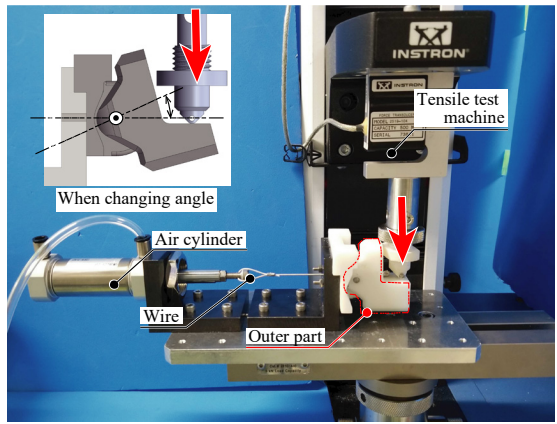

Figure 11. Experimental setup for holding torque

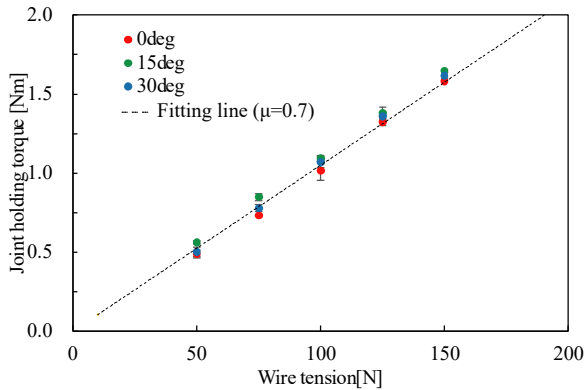

Figure 12. Experimental result for holding torque by friction

The measured frictional holding torque in bending direction is shown in Fig. 12. The relationship between the wire tension and the holding torque was linear. The holding torque was smaller when the joint angle was greater than 0° than when the joint angle was 0°. The fitting curve obtained from Eq. (2) for a friction coefficient of 0.7 is denoted by the dashed line.

*B. Payload Measurement of Proposed Structure*

The stiffness of the proposed mechanism to support weight was measured using an experimental machine to determine the tendency of the stiffness to increase with internal air pressure. For the experiment, we used the integrated model shown in Fig. 10. The amount of deflection was measured with two weights, 150 g and 500 g, attached 400 mm from the base. Then, the displacement was measured when a pressure of 0–10 kPa was applied. Measurements were recorded ten times, and the average of value was used for the results. In addition, experiments were conducted with 150 N of jamming wire tension applied and with no skeleton.

Fig. 13 shows the relationship between the internal air pressure and the measured amount of deflection. The measurement results are evaluated in two different ways. In the proposed mechanism, deflection occurs before the weight is attached due to the weight of the skeleton. In fig. 13(a), the displacement from the straight position (position when gravity is zero) was plotted against pressure and the mass of the attached weight. In fig. 13(b), the displacement from the initial position which structure is deflected by its own weight was plotted. In both graphs, dashed line, chain line, and full line show the result of beam composed only membrane, membrane with skeleton, and membrane with wire jammed skeleton. In addition, the red lines indicated attached 500g weight, and blue lines showed one of attached 150g weight. However, the result of no skeleton with 500g weight be included in graph (b) because the deflection was too large.

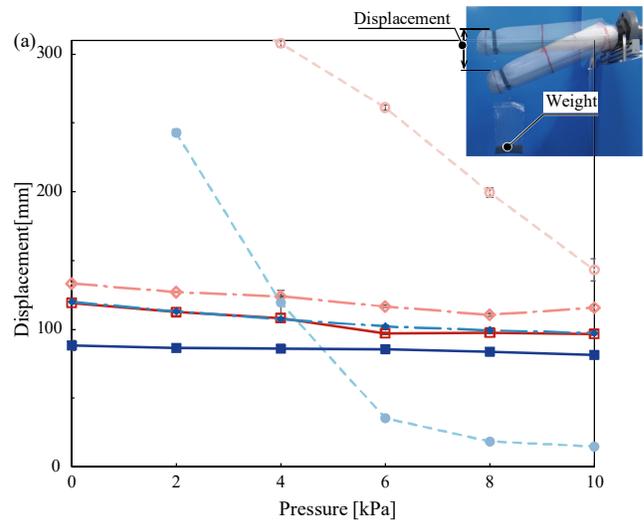

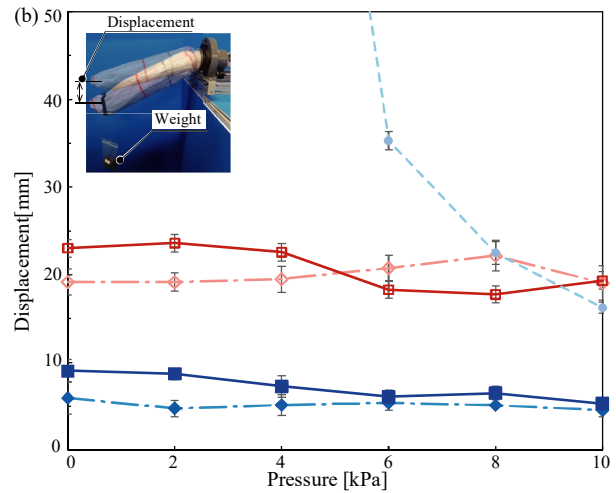

Figure 13. Relationship between wire tension and displacement.
(a) Displacement from the straight position, (b)Displacement from the initial position which structure is deflected by its own weight

*C. Movement of Prototype*

We conducted an operation test using an integrated model that bends to the shape of the human body and retains its shape to support raising the human's upper body (Fig. 14), as follows. First, the internal space of the membrane structure was pressurized to extend with inserting between arm and torso. Subsequently, the bending wire inside the articulated links was pulled to bend and adhere to the body while keeping compliant with pressurized. The central wire was then pulled to increase the contact force and fix the shape. After fixing the shape, arise the human's upper body by moving the entire mechanism by operated in hand. For the retraction of the membrane to its original state, it was first pressurized and then bent back to a straight shape using the side wires of the articulated links in a direction opposite to that when it was attached to the human body. Next, the wire was rewound from the end of the articulated links so that it would return to the containment. The motor then rewound the retraction wire attached to the end of the articulated link (see supplementary video).

## V. Discussion

### A. Discussion of Experimental Results

In this study, the load-carrying capacity of the tip-extension mechanism with an articulated link structure was experimentally measured and evaluated. In this section, we discuss the types of design improvements that are necessary for situations where this mechanism can be applied. First, in the measurement of the holding torque by friction, as described in Section IV A, holding torque of approximately 1.6 Nm can be generated by a maximum wire traction force of 150 N. This means that a force of approximately 3.2 N can be retained when the wire is extended by a maximum length of 500 mm. With this holding force, it is difficult to support a heavy object, such as a human body. For such objects, improvement measures can be considered, such as increasing the radius of the link where friction acts, using a material with higher friction, or by dividing the contact surface into multiple layers to increase the contact surface [23]. In addition, tension can be added to the bending wires to generate bending torque within a range that does not exceed the holding torque due to friction, thereby further improving the holding torque. In contrast, the bending torque can be decreased for application of the mechanism in situations where it is sufficient to support its own weight, such as in search applications.

The measured deflection in the direction of gravity shown in Fig. 13(a) and (b) indicate that the stiffness can be maintained at low pressure by using the articulated links. In graph (a), the deflection was reduced by only about 15% to 20% by applying 10 kPa of pressure to the proposed mechanism without pressure. At the low pressure of 10 kPa or less, the deflection can be reduced for a light load of 150 g without the link. On the other hand, for a large load of 500g, the frame could not hold the object because of the large deflection. It is known that low pressure causes wrinkles near the base side of the Inflatable beam. High pressure is required for the membrane-only structure not only to reduce the deflection but also to prevent the beam from breaking. Stiffening by 150N wire tension reduced the deflection by an average of about 13.3% (16mm) at 500g and 19.6% (21.2mm) at 150g. Comparing Fig. 13(a) and (b), it can be seen that most of the deflection is due to the weight of the articulated links, since the deflection when the weight is placed is 10 mm to 20 mm, while the deflection when the links and weight are loaded together is around 100 mm. The deflection due to its own weight is caused by the clearances in the joints, and it will be necessary to adjust the clearances or reduce the weight of the links. It is possible to increase stiffness by applying larger pressure. However, the contact force becomes larger, and compliance is lost. Also, the tension required for retraction is also larger, and the buckling occurs during the retraction process.

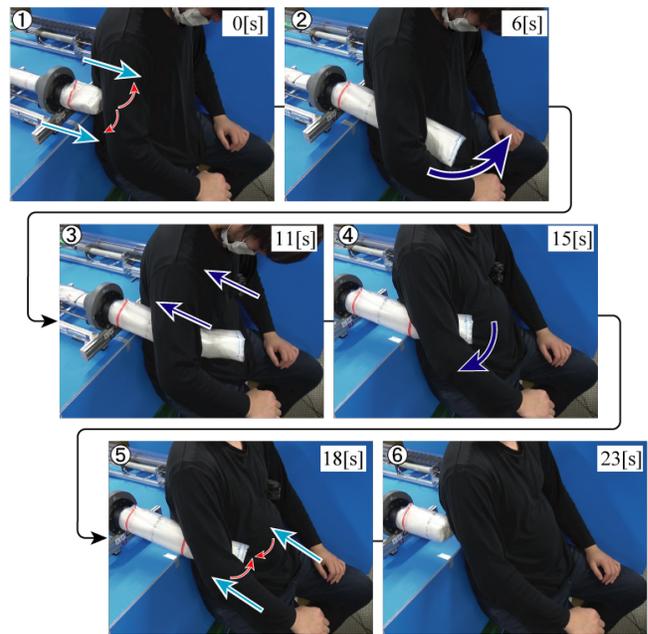

Figure 14. The movement of bending for human body

### B. Limitation of proposed mechanism

The proposed mechanism, in combination with a rigid link structure, improves the load-bearing capacity and expands the range of adaptable situations; however, it has some limitations. First, the proposed mechanism can only move in the same plane along the Z-axis direction. It is also possible to turn it upward by rotating the entire structure in a torsional direction after curving it; however, in this case, the range of motion is limited. In addition, the insertion of articulated links limits the narrow space that can be passed through. However, these articulated links can be designed to be smaller depending on the application. Link insertion may interfere with the extending operation of the eversion structure, but this is not a major problem. In the configuration in this paper, the mechanism can extend by applying low pressure (at least 4kPa).

## VI. Conclusion

In this study, an articulated link structure made of a hard material was inserted inside a tip-extension structure, which was composed of a thin membrane, to achieve both high load-bearing capacity in the direction of gravity and low frictional insertion, which are the advantages of the tip-extension structure. In the proposed mechanism, an articulated link with two axes of rotation, the pitch axis, and the yaw axis, was used as the articulated link structure for the tip-extension movement of the membrane. The range of motion was restricted in the pitch axis; therefore, the robot could support its own weight. When wires are passed through the articulated links, the robot can bend and maintain its shape in the yaw direction.

The articulated links must be both lighter and more rigid. This will reduce the effect of deflection caused by the weight of the skeleton, as determined from the experiments. For this purpose, a material with a high strength-to-mass ratio, such as aluminum alloy (Fig. 15(a)), should be used. To improve rigidity and reduce weight, it is not necessary to change the material or design for the entire articulated link; only the part near the base, where larger moments due to the weight are applied, can be considered.

In the proposed mechanism, the articulated link lies along a straight line, which increases the overall size of the device. However, the proposed mechanism can be rolled up and stored, as shown in Fig. 15(b). For example, the 970 mm skeleton in the proposed mechanism can be stored in a configuration with a diameter of about 180 mm.

APPENDIX

In this section, the picture illustrating future improvements to the mechanism is shown below.

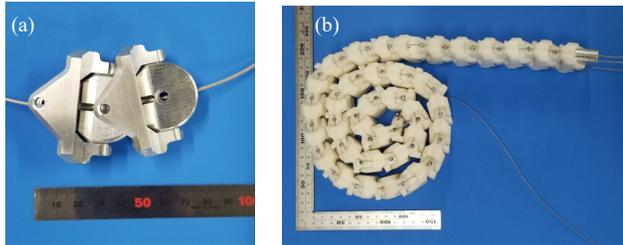

Fig. 15 (a) Assembly of link parts made by aluminum alloy
(b) Articulated links stored by reeling